\documentclass[letterpaper, 10pt, conference]{ieeeconf}      
\IEEEoverridecommandlockouts                              

\usepackage{graphicx} 
\usepackage{times} 
\usepackage{amsmath} 
\usepackage{bm}
\usepackage{ragged2e}
\usepackage{xfrac}
\usepackage{amssymb} 
\usepackage{algorithm}
\usepackage[noend]{algpseudocode}
\usepackage{multirow,color}
\usepackage{cite}
\usepackage{algpseudocode}
\usepackage{varwidth}
\usepackage{subfig}
\usepackage{booktabs}
\usepackage{breqn}
\usepackage{gensymb}
\usepackage[hyphens]{url}
\usepackage[export]{adjustbox}
\usepackage[font=footnotesize]{caption}
\let\labelindent\relax
\usepackage[inline]{enumitem} 
\usepackage{dblfloatfix}
\usepackage{tabularx}
\usepackage{setspace}
\usepackage{textcomp}
\usepackage{multirow}
\usepackage{comment}
\usepackage{xcolor}
\usepackage[export]{adjustbox}

\usepackage{wrapfig}

\usepackage{color}
\usepackage{graphicx,tipa}

\makeatletter
\def\BState{\State\hskip-\ALG@thistlm}
\makeatother

\title{\LARGE \bf Towards a Novel Wearable Robotic Vest for Hemorrhage Suppression}

\author{Harshith Jella$^*$, Pejman Kheradmand$^*$, Joseph Klein, Behnam Moradkhani, and Yash Chitalia
\thanks{\textit{$^*$Harshith Jella and Pejman Kheradmand contributed equally to this
work.}}
\thanks{All authors are with the Healthcare Robotics and Telesurgery (HeaRT) Laboratory, Speed School of Engineering, University of Louisville, Louisville, KY, USA.}
\thanks{\textit{Corresponding author: Pejman Kheradmand} (pejman.kheradmand@louisville.edu)}
}
\begin{document}
 \maketitle
\begin{abstract}
This paper introduces a novel robotic system designed to manage severe bleeding in emergency scenarios, including unique environments like space stations. The robot features a shape-adjustable "ring mechanism", transitioning from a circular to an elliptical configuration to adjust wound coverage across various anatomical regions. We developed various arms for this ring mechanism with varying flexibilities to improve adaptability when applied to non-extremities of the body (abdomen, back, neck, etc.). To apply equal and constant pressure across the wound, we developed an inflatable ring and airbag balloon that are compatible with this shape-changing ring mechanism. A series of experiments focused on evaluating various ring arm configurations to characterize their bending stiffness. Subsequent experiments measured the force exerted by the airbag balloon system using a digital scale. Despite its promising performance, certain limitations related to coverage area are identified. The shape-changing effect of the device is limited to scenarios involving partially inflated or deflated airbag balloons, and cannot fully conform to complex anatomical regions. Finally, the device was tested on casualty simulation kits, where it successfully demonstrated its ability to control simulated bleeding. 
\end{abstract}

 

\section{Introduction}\label{sec:intro}
When considering space travel and the potential for injuries, there are significant differences in how treatment is handled in space compared to on Earth. While there have been no reported cases of major hemorrhages in space so far, during longer missions, such as a journey to Mars, this could be life-threatening \cite{pantalone2022facing, kirkpatrick2001extraterrestrial}. Although the chances of severe trauma or a surgical emergency are low, concerns about crushing injuries, penetrating trauma, or other conditions requiring surgery remain significant. The differences in treatment are primarily attributed to the lack of gravity. In the absence of gravity, there are certain physiological changes, such as reduced circulation in the feet and increased circulation in the head and torso \cite{pantalone2022facing}, and as a result, differences in how one approaches the situation. On Earth, when one is bleeding heavily the immediate instinct is to apply pressure to the wound, which is feasible in both extremities and non-extremities thanks to gravity. However, in space, this poses several challenges. The most significant challenge is that both the injured person and the responder must brace themselves against some external surface to apply the necessary pressure. 

\begin{figure}[t]
\vspace{4 mm}
\centering\includegraphics[width=\linewidth,keepaspectratio]{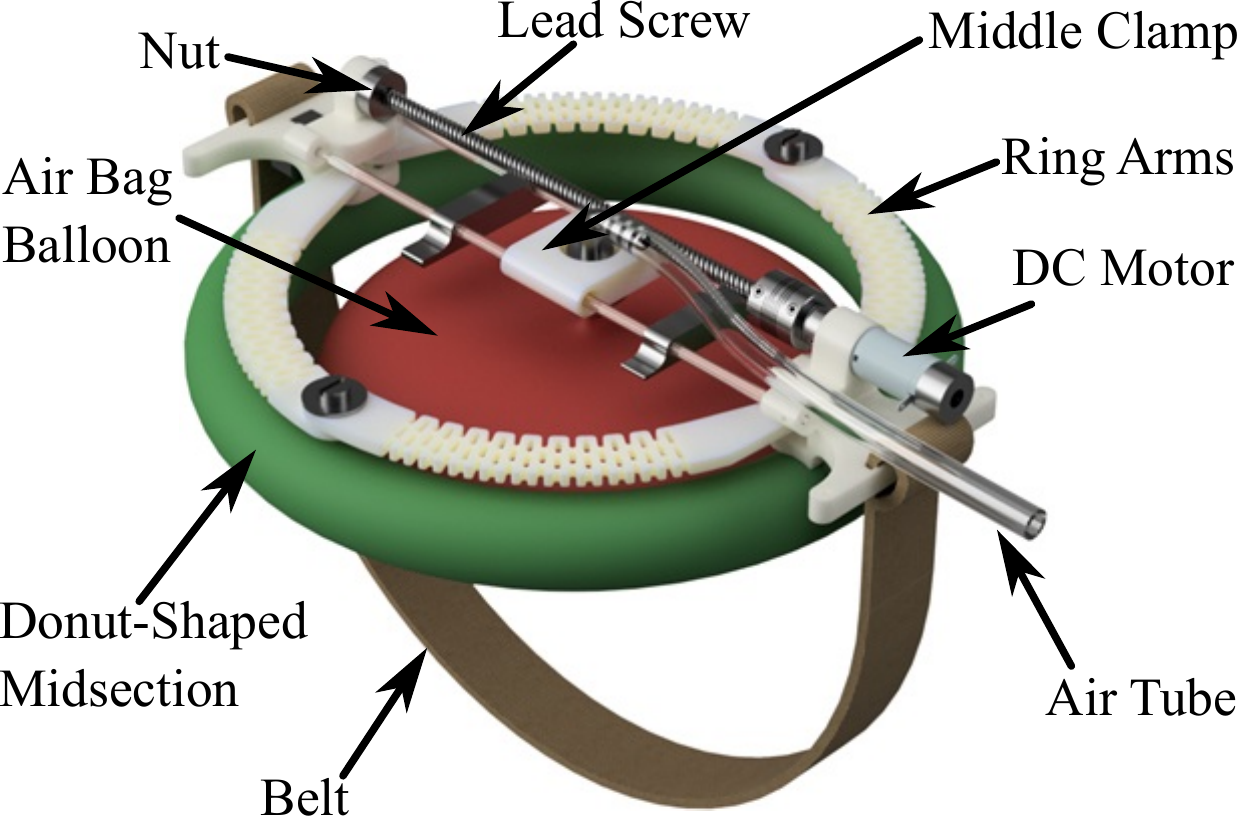}
\caption{Overall wearable robot design with labeled components.}
\label{fig:designfig}
\vspace{-5 mm}
\end{figure}

Current protocols for managing hemorrhage in outer space focus on temporarily stabilizing the patient until they can be returned to Earth for definitive treatment \cite{pantalone2022facing}. However, this approach may not be feasible during extended space missions. Existing hemorrhage control devices have been proven to be highly effective at controlling hemorrhage and improving survival rates within terrestrial pre-hospital settings \cite{shi2021prehospital}. However, their use is primarily limited to the extremities and require proper training \cite{thoolen2023external}. For non-extremity (axial skeleton and torso) hemorrhaging, the available options include: Resuscitative Endovascular Balloon Occlusion of the Aorta (REBOA), and abdominal aortic and junctional tourniquet (AAJT)\cite{zhang2024torso}. However, the use of either devices come with several potential complications\cite{zhang2024torso} due to their mechanism of action - blocking off the aorta, the main artery that supplies blood from the heart to the entire body. Stopping hemorrhage using REBOA requires a person guiding a catheter up the femoral artery to the aorta under the guidance of an x-ray and inflating a small balloon which blocks the flow of blood. The complicated nature of this procedure requires personnel extensive training and may result in a variety of side effects including: acute kidney injury, sepsis and pulmonary issues. On the other hand, the AAJT stops hemorrhage by applying external pressure on the trunk of the body and compressing the aorta. As a result, the pneumatic portion of the belt can exert excessive pressure on surrounding tissues, which has been associated with tissue damage, including necrosis in the liver and intestines, and ischemia. While user-friendly, the device is not fully optimized for use in space \cite{zhang2024torso}. Devices that are sent to outer space have to conform to a set of constraints specifically in relation to the size and weight of the object. REBOA catheters are very small but require chest X-rays, while traditional AAJT devices are bulky, and require a hand pump. Furthermore, both devices are one-time use. The AAJT also lacks a precise method for measuring the pressure being applied, offering only a basic gauge for estimation. Additional hemostatic methods include adhesives, mechanical hemostats, sealants, and hemostatic dressings. These are difficult to apply in an emergency in a space vehicle due to the lack of a large set of personnel. If only one responder is present on-board, this responder cannot relieve the applied pressure on the laceration to apply the aforementioned hemostatic dressings. Therefore, an adaptable automated solution is needed to apply constant pressure on non-extremity wounds, that can adapt to the shape of the laceration. 
\begin{figure}[]
\centering\includegraphics[width=\linewidth,keepaspectratio]{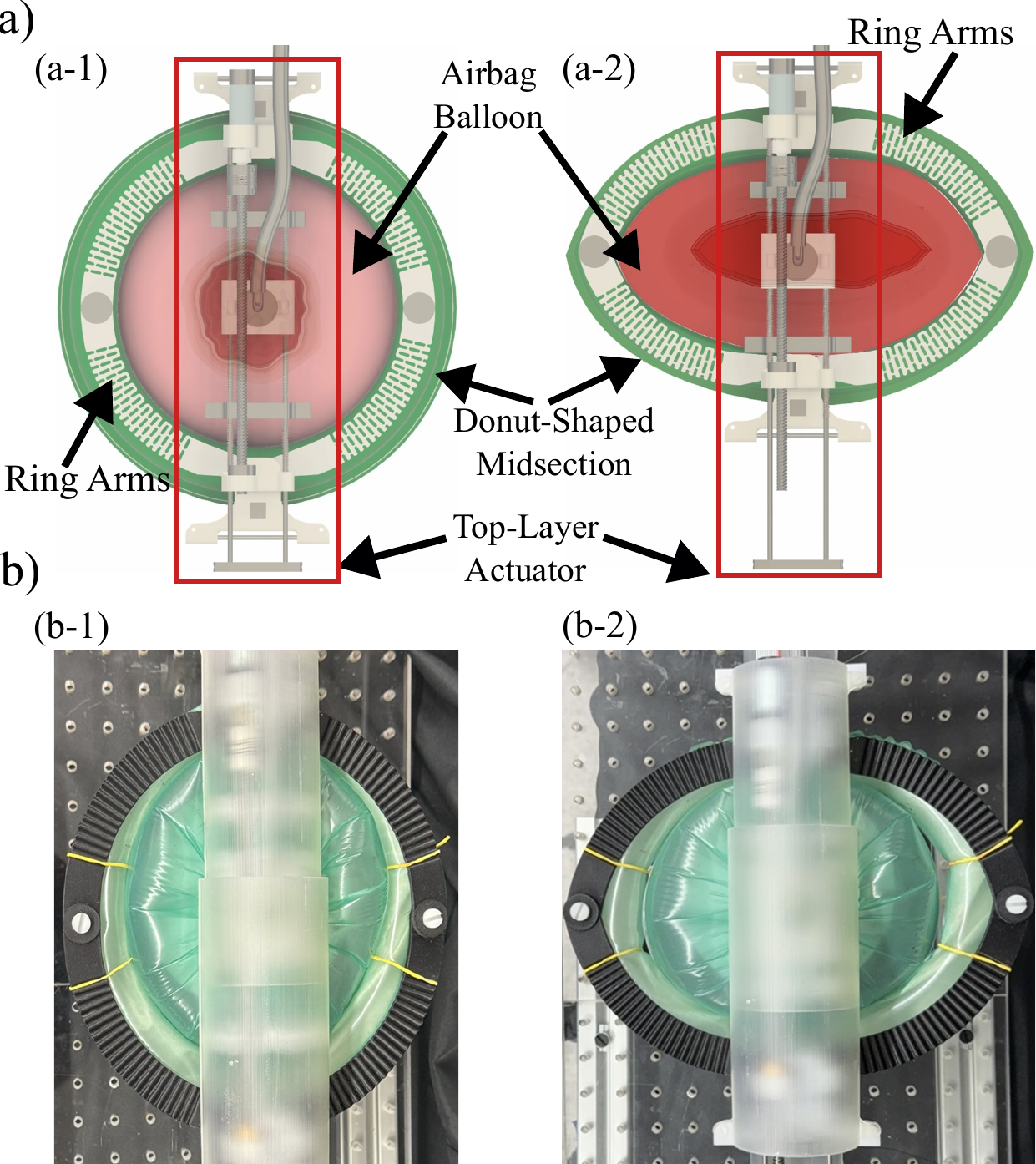}
\caption{Configurations of the wearable robot, (a-1) Design model in the circular configuration for addressing circular wound, (a-2) Design model in the elliptical configuration for linear wound, (b-1) Actual implementation of the robot in the circular configuration, (b-2) Actual implementation of the robot in the elliptical configuration.}
\label{fig:Configuration}
\vspace{-6 mm}
\end{figure}

Designing such an automated device that interacts with humans requires consideration of several factors. Conventional robotic systems in healthcare, with their rigid arms and bodies, have limited flexibility and can be unsafe for human interaction. In contrast, soft robots are becoming more popular because they are flexible, lightweight, and inexpensive. They are also safer in environments with humans as their materials can absorb energy from collisions, reducing the risk of harm\cite{Rus2015}. Due to these advantages, soft robots have been utilized in a range of applications, including jamming and gripping mechanisms\cite{soro20150019}, worm-like robots\cite{MagWorm}, vine robots\cite{VineRobots}, and surgical robots\cite{A_Collapsible_Soft_Actuator}. A specific fabrication method for soft robots involves using a heat press to create robots from thin layers of plastic, which are bonded together through a heat-sealing process\cite{LayeredManufacturing}. In this study, we present the wearable robot to  apply constant pressure to control hemorrhaging in non-extremity wounds. The proposed device offers several contributions to the field. It introduces a novel design with an actuated ring arm mechanism capable of adapting the shape of the compression region, optimizing tissue contact, and localizing the applied pressure effectively to address the variability in wound geometries (see Figs. \ref{fig:designfig} and \ref{fig:Configuration}). The pressure is applied through an inflatable ring and balloon mechanism, ensuring consistent and controlled force tailored to the wound area. Furthermore, the design emphasizes reusability and practicality for remote applications, with inflatable components that can be replaced after each use while keeping the mechatronic housing reusable, as it does not come into direct contact with the wound. Finally, the device builds on the existing FDA-approved AAJT, integrating advanced robotic elements to enhance adaptability, usability, and overall patient outcomes.

The remainder of this manuscript is organized as follows: In Section \ref{sec:design}, we describe the device's design, including the top layer actuator (\ref{subsec:Top Layer Actuator}), the donut-shaped mechanism (\ref{subsec:Inflatable Ring}), and the airbag balloon (\ref{subsec:Airbag Balloon}). Section \ref{sec:expRes} focuses on modeling and experiments, where we model the ring arm's flexibility (\ref{subsec:mechanical_model}), test different ring arm designs (\ref{subsec:RingArmsec}), measure the maximum pressure tolerable by inflatable elements (\ref{subsec:BurstPressureExperiment}), and evaluate the pressure and force the device applies to surfaces (\ref{subsec:test_contact_surface}). In Section \ref{subsec:test_casualty_simulation}, we test the robot on a casualty simulation to assess its efficacy in hemorrhage control. Finally, in Section \ref{sec:conclusions}, we summarize the project and outline future directions.

\begin{figure}[t]
\centering\includegraphics[width=\linewidth,keepaspectratio]{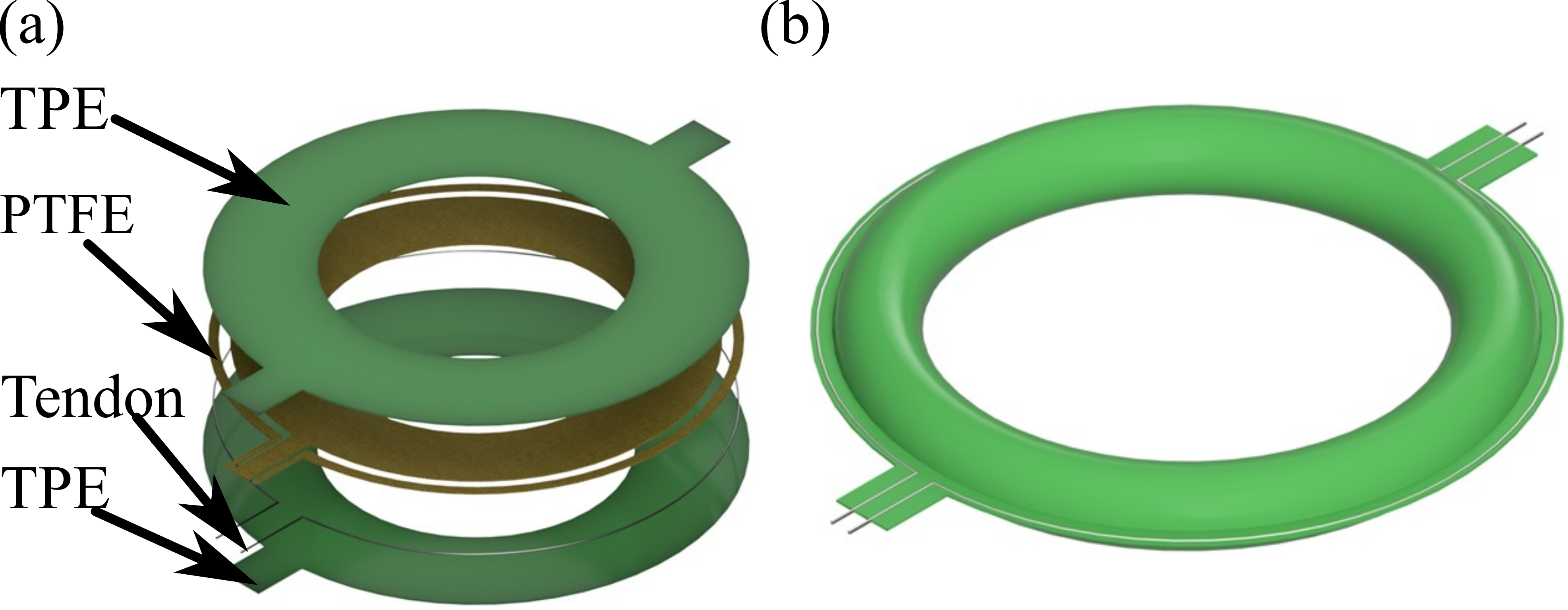}
\caption{Midsection ring, (a) a layered schematic illustrating the order of each layer (b) the inflated ring configuration. }
\label{fig:outer_ring}
\vspace{-2 mm}
\end{figure}

\section{Robot Design} \label{sec:design}
Our wearable device is composed of three different components: a rigid top layer that facilitates manual shape adjustment to match the target laceration’s contours, a soft and inflatable ring-shaped midsection, and a central airbag balloon designed to apply direct pressure to the wound. 

\subsection{Top Layer Actuator}\label{subsec:Top Layer Actuator}
The top layer features a DC motor ($\phi16$~mm Precious Metal Brushes CLL, 1.2 Watt) connected to a high-speed lead-screw (1018 carbon steel, ${1/4}^{"}$ – 12 thread size, McMaster-Carr). One end of the device houses the motor, while the other houses the nut. Each end is attached to two ring arms, which link together at the center of the device to form a hinge (See Fig. \ref{fig:designfig}). Upon motor activation, the lead-screw rotates, moving the attached nut along its length which shifts the shape of the ring formed by the ring arms from circular (See Fig. \ref{fig:Configuration} (a-1)) to elliptical (See Fig. \ref{fig:Configuration} (a-2)). The ring arms are also specifically designed to flex and adjust their curvature, conforming to the body shape as the robot wraps around it. Three different ring arms were designed: standard, cutouts, and ridges, which are discussed in detail in Section \ref{subsec:RingArmsec}. The standard ring arms are a solid arm with a constant thickness of $4$~mm and no cutouts. The cutout ring arms are a $4$~mm thick arm with elliptical cutouts throughout its midsection. The ridges ring arms are $4$~mm thick where they interface with other parts and have a thinner $1.5$~mm ridge throughout its midsection. In the center of the actuation mechanism, there are two brackets which serve as a clamp to mount the airbag balloon to the device (See Fig. \ref{fig:designfig}). It is supported by two stainless steel elements and four springs on either side to ensure consistent central alignment.

\subsection{Inflatable Ring}\label{subsec:Inflatable Ring}
The midsection primarily consists of an inflatable ring that constrains the edges of the airbag balloon, so when the balloon inflates, its expansion is limited within this ring. With the ideology of this being single use and also constrained for space, this ring was made using a layered manufacturing method using thermoplastic elastomer (TPE), and nonstick polytetrafluoroethylene (PTFE) layers \cite{A_Collapsible_Soft_Actuator}. The PTFE is sandwiched between two layers of TPE and heat pressed for four minutes at $250\degree$ to create a strong bond between the TPE in areas uncovered by the PTFE as illustrated in Figure \ref{fig:outer_ring}. There is a large PTFE ring in the middle to make an inflatable segment, and a thin strip around the circumference of the TPE. Two tendons are routed through the thin strip along the ring's edges and are anchored to a small bracket that allows for it to be attached to the end bracket (See Fig. \ref{fig:outer_ring} (a)). Thus, when the motor actuates and reshapes the top layer, the ring's shape adjusts accordingly. The overall thickness of the inflatable ring is $0.101$~mm and takes up $1788$~mm$^{3}$ when deflated and expands to an overall volume of $8350$~mm$^{3}$ at $4.83$~Kpa of pressure.
\begin{figure}[t]
\centering\includegraphics[width=\linewidth,keepaspectratio]{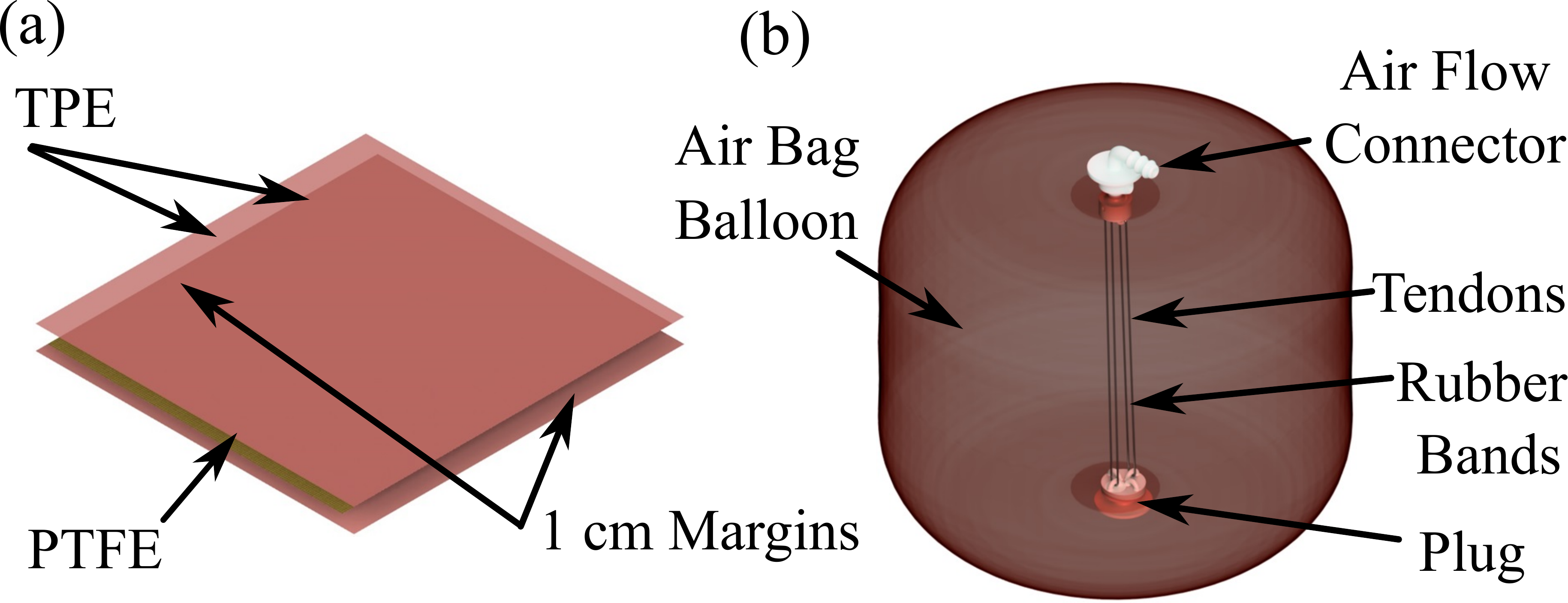}
\caption{The airbag balloon, (a) a layered schematic illustrating the different layers used, (b) translucent inflated airbag balloon showing all the internal and external components.}
\label{fig:ballooninflated}
\vspace{-4 mm}
\end{figure}

\subsection{Airbag Balloon}\label{subsec:Airbag Balloon}
The final component of this device is the airbag balloon, which is mounted onto the aforementioned middle bracket. The balloon is manufactured from two rectangular sheets of TPE with a $1$~cm heat pressed margin on two parallel ends (See Fig. \ref{fig:ballooninflated} (a)). There are two different brackets with which the TPE interfaces, an air flow connector and a lower plug (See Fig. \ref{fig:ballooninflated} (b)). Each open end of the airbag balloon is rubber banded onto these brackets. These brackets also allow for the attachment of internal rubber bands and tendons (see Fig \ref{fig:ballooninflated}). The rubber band allows for the balloon to remain compact and stable when deflated, preventing flimsiness, while the tendons limit the max extension of the airbag balloon. Without the tendons, the balloon overextends and the rubber band becomes undone. Upon inflation, the balloon expands to apply controlled pressure to the laceration area. The airbag balloon is $0.076$~mm thick and takes up  $13,169$~mm$^{3}$ when deflated and expands to an overall volume of $1,830,508$~mm$^{3}$
 at $4.83$~Kpa of pressure.

\section{Mechanical Modeling, Experiments and Results}\label{sec:expRes}
This section outlines a mechanical model for the ring arm along with four experiments conducted to evaluate the robot's performance: (1) measurement of the bending stiffness of the ring arm component, (2) evaluation of the burst pressure to measure the maximum pressure that can be applied to the airbag balloon and ring, (3) validation of the robot's ability to apply the desired pressure to a surface, and (4) assessment of the robot's effectiveness in stopping bleeding using a human torso model with simulated blood.

\begin{figure}[t]
\centering\includegraphics[width=\linewidth,keepaspectratio]{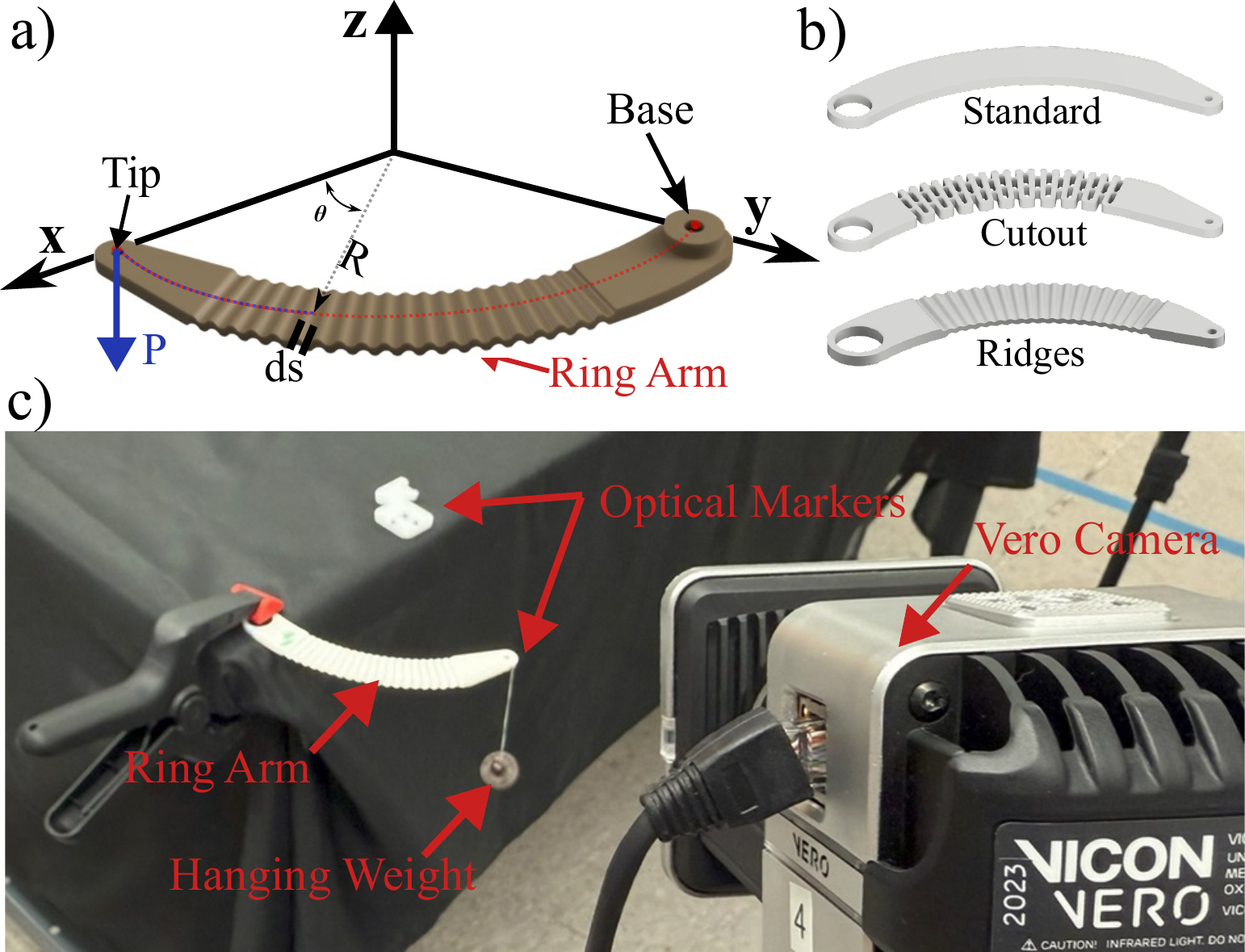}
c\caption{(a) The ring arm shown as an arc-shaped curve (in $x-y$ plane) subjected to the perpendicular force $P$ (in $-z$ direction) at its tip ($x=R$) and mechanically secured at the base ($y=R$), (b) the 3 characterized ring arms: Standard, Cutout, and Ridges, (c) the isolated ring arm experimental setup.}
\label{fig:castiglinao}
\end{figure}

\begin{table}[t]
\centering
\caption{Ring arm bending stiffness results}
\label{tab:bending}
    \begin{tabular}{|c|c|c|c|}
    \hline
         Design type & Standard&Cutout&Ridges\\
         \hline
         $EI (N.m^2)$x$10^{-7}$& 8.9$\pm$0.66&
         3.2$\pm$0.04&
         2.7$\pm$0.03\\
         \hline
    \end{tabular}
\vspace{-3 mm}
\end{table}

\subsection{Mechanical Model}\label{subsec:mechanical_model}
When the robot is placed on a laceration, the top layer actuation system (described above in Section \ref{subsec:Top Layer Actuator}) positions the ring arms such that the shape of the airbag balloon closely mimics the nature of the wound. The arms then rotate about the fixed hinges at either end to change the shape of the airbag. Therefore, the ring-arms are to be actuated in-plane with the natural curvature of the arms (i.e. the $x$-$y$ plane in Fig. \ref{fig:castiglinao}(a)). When placed on a non-extremity, however, the arms are likely to undergo flexion in a direction orthogonal to this in-plane bending (see $z$-axis in Fig. \ref{fig:castiglinao}(a)) due to the contours of the human torso. It is therefore essential to design ring arms that allow for this deformation without permanent damage to the Top Layer Actuator mechanism. 
Since the ring arm undergoes undesired out-of-plane bending, Castigliano's first theorem was applied to model its deflections and calculate the bending stiffness\cite{boresi1993advanced}. The design with the lowest bending stiffness was selected for this preliminary design. In our future work, we hope to conduct a formal optimization of the ring arm geometric parameters for this out-of-plane bending.

For simplicity, the ring arm component was approximated as a quarter-circle with a specified width. The primary strain energy was assumed to result from bending, while the effects of axial, shear, and twist strains were disregarded. The ring-arms are modeled as pre-curved cantilever beams fixed at their base, and subjected to tip forces (from the hinge reaction forces at the tip, indicated by $P$ in Fig. \ref{fig:castiglinao}(a)). By applying Castigliano's theorem, the tip deflection of the ring arm is determined as:
\begin{equation} \label{eq:castigliano formula}
    q_P =\int_{s=0}^{L} \frac{M}{EI} \frac{\partial M}{\partial P} \,ds
\end{equation}
where $q_P$ is the ring arm's tip deflection due to the exerted out-of-plane force $P$, $M$ is the moment produced by the force $P$ along the ring arm, $s$ is the arc length defined on the ring arm (see Fig. \ref{fig:castiglinao}(a)), and finally $EI$ is the constant beam stiffness ($E$ and $I$ are the beam's stiffness modulus and the beam's area moment of inertia respectively). Using the assumption stated above, the moment can be written as $M=PR\sin(\theta)$, with $R$ being the radius of the ring arm arc-shaped curve, and $\theta$ being the defined arc angle (see Fig. \ref{fig:castiglinao}(a)). Therefore, the partial derivative of the moment with respect to the perpendicular force can be calculated.

\begin{equation} \label{eq:partial derivative}
    \frac{\partial M}{\partial P} = R\sin(\theta)
\end{equation}
Substituting the obtained partial derivative in Eq. (\ref{eq:partial derivative}) into the general Castiglinao's formula, represented in Eq. (\ref{eq:castigliano formula}), and replacing moment and arc length with their corresponding equivalent expressions in our case, the integral can be written as:
\begin{equation}
    q_P = \int_{0}^{\frac{\pi}{2}} (\frac{PR\sin(\theta)}{EI}) (R\sin(\theta)) R\,d\theta
\end{equation}
Rearranging the obtained integral equation gives us the expression for the ring arm's bending stiffness in terms of the perpendicular force and tip deflection:
\begin{equation}
   EI= \frac{\pi}{4} \frac{PR^3}{q_P}
\end{equation}

 \subsection{Ring Arm Experiment}\label{subsec:RingArmsec}
\begin{figure}[t]
\centering\includegraphics[width=\linewidth,keepaspectratio]{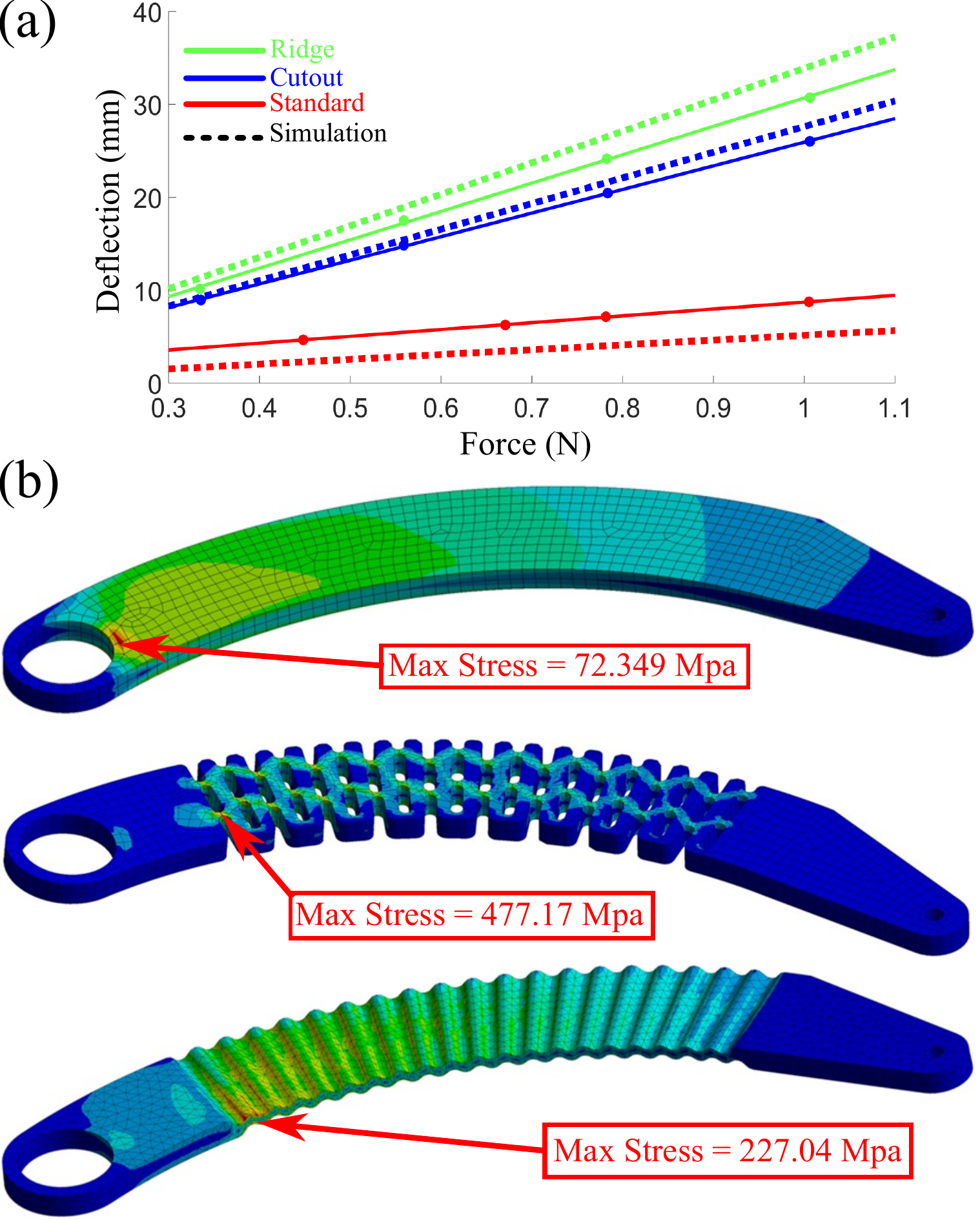}
\caption{(a) Deflection at the tip of the ring arm as a function of applied force, with experimental results compared to simulations performed in ANSYS. (b) The stress distribution results obtained from ANSYS simulations for a $20$~N applied force, showing the maximum stress values for each design.}
\vspace{-9 mm}
\label{fig:EI_matlab}
\end{figure}

 To evaluate the bending stiffness of the three different ring arm designs (see Fig. \ref{fig:castiglinao}(b)), we developed an experimental setup using four Vero v2.2 motion tracking cameras (Vicon Motion Systems Ltd., United Kingdom) to monitor a marker placed at the tip of each ring arm (see Fig. \ref{fig:castiglinao}(c)). We applied four distinct loads to each arm, and used the cameras to accurately measure the resulting displacement at the tip for each load. Figure \ref{fig:EI_matlab}(a) illustrates the applied force at the ring arm tip versus the deflection. Each data point represents a specific applied force, and a fitted line through these points confirms that the arm operates within the elastic region. ANSYS simulations were conducted for each design, applying a force to the tip and calculating the resulting deflection, which was then compared to the experimental results. The graph demonstrates close agreement between the experimental data and simulation predictions, validating the accuracy of the computational model. Table \ref{tab:bending} summarizes the bending stiffness evaluation for the tested ring arm designs. We also analyzed the maximum stress for each design under a $20$~N force (see Fig. \ref{fig:EI_matlab}(b)). The results reveal that while the standard design experienced the lowest maximum stress, it exhibited limited flexibility. The cutout design displayed the highest maximum stress, with stress concentrations around the cutout section, indicating a potential failure point under load. The ridge design emerged as the best choice among these broad designs, offering greater flexibility and better stress distribution under the same applied force, making it both durable and adaptable. As mentioned previously, a future goal is to further optimize the ridges to reduce maximum stresses on the ridges while retaining flexibility in the ring arm.

\subsection{Burst Pressure experiment}\label{subsec:BurstPressureExperiment}
To test the burst pressure of the Inflatable Ring (see Section \ref{subsec:Inflatable Ring}) and the Airbag Balloon (see Section \ref{subsec:Airbag Balloon}), both components were slowly inflated using the airflow manifold and the pressure was read using the pressure sensors. The burst pressure of the ring was determined to be $16.55$~Kpa, while the burst pressure of the balloon was $18.62$~Kpa. The primary failure point of the ring is the seal on the inside edge of the ring - there were inconsistencies in the thickness of the sealed margin due to  a lack of positioning brackets for the TPE and PTFE. However, the failure of the balloon was not due to the seal, but primarily due to the attachment point between the TPE and the plug. Because the TPE was simply attached with a rubber band, small segments of the TPE would prone to slipping and releasing the pressure.

\begin{figure}[t!]
\centering\includegraphics[width=\linewidth,keepaspectratio]{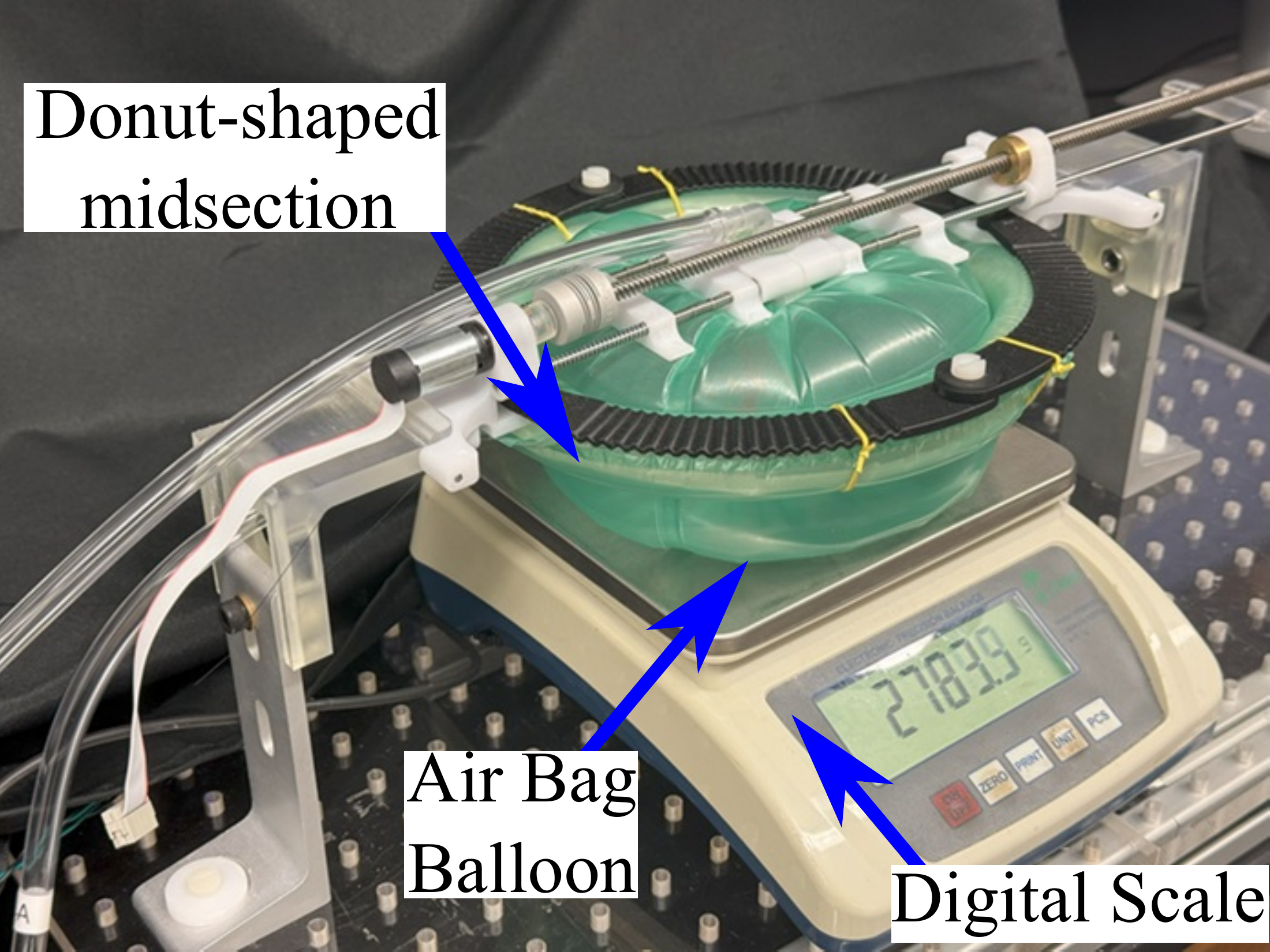}
\caption{Pressure testing setup illustrating the robot applying force to a surface, with the applied force measured using a digital scale positioned beneath the airbag balloon.}
\label{fig:TestFig}
\end{figure}

\begin{figure}[]
\centering\includegraphics[width=\linewidth,keepaspectratio]{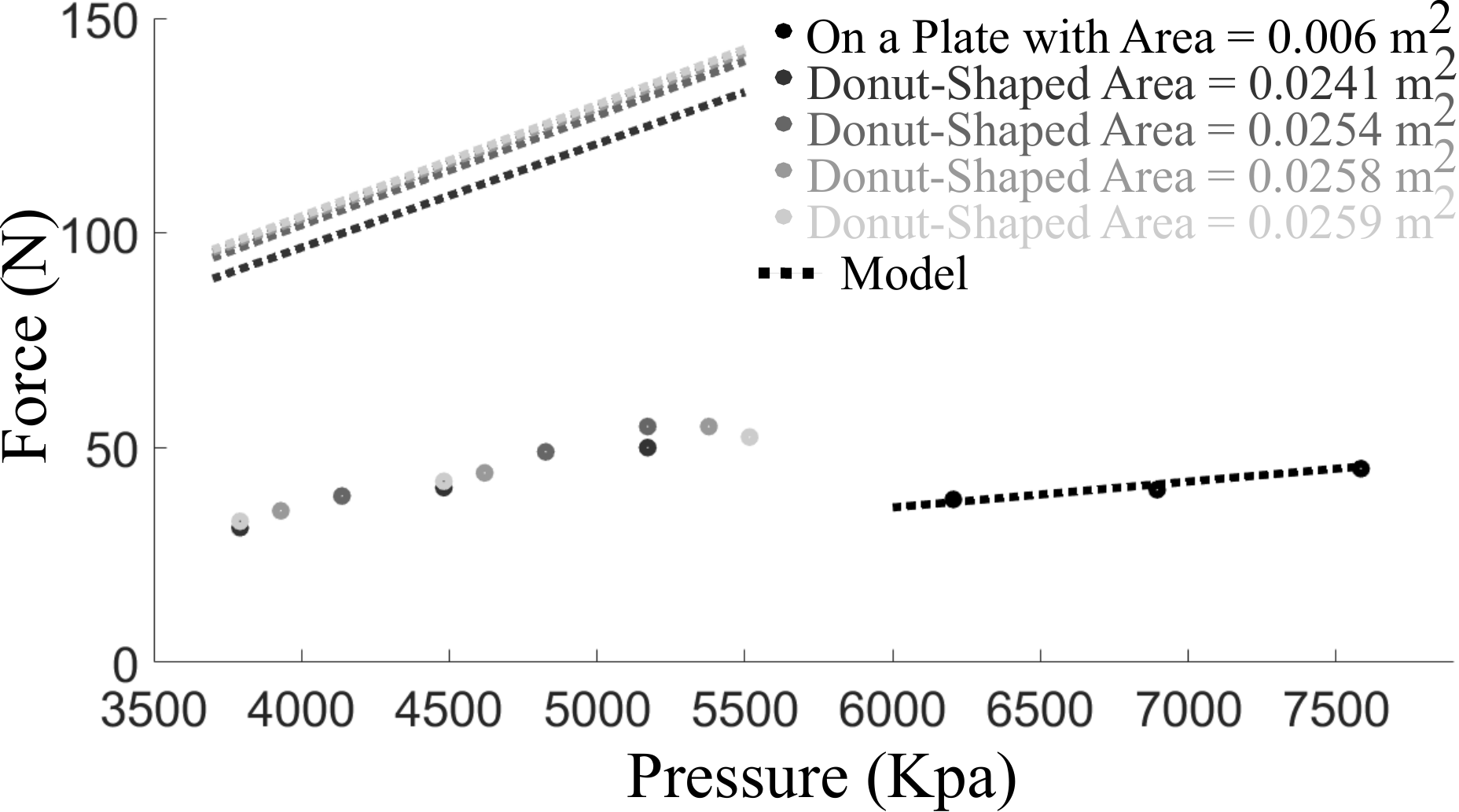}
\caption{Comparison of the force exerted on the contact surface versus the pressure applied to the airbag balloon, showing both theoretical and experimental results.}
\label{fig:Testresult}
\vspace{-8 mm}
\end{figure}

\subsection{Test on contact surface }\label{subsec:test_contact_surface}
A validation setup was developed to measure the pressure exerted by the robot on the body. In this setup, the robot is securely mounted on an acrylic breadboard, and a digital scale is positioned beneath the airbag balloon. As the balloon is inflated, it expands and applies force against the digital scale (see Fig. \ref{fig:TestFig}). The force applied to the contact surface can be estimated using the equation:
\begin{align}\label{eq:forcetest}
    F = A \times (P_{\text{balloon}} - P_{\text{atm}})
\end{align}
where $A$ represents the contact area, $P_{\text{balloon}}$ is the regulated input pressure controlled by a closed-loop pressure regulator (QBX series, Proportion-Air), and $P_{\text{atm}}$ is atmospheric pressure. Various experiments were conducted using different shapes for the donut-shaped midsection, including circular and oval configurations with varying major axes. Figure \ref{fig:Testresult} illustrates the relationship between the force applied to the digital scale and the pressure in the airbag balloon. The data points represent the experimental data, while the fitted lines indicate the predicted force applied to the surface based on the pressure input. Four experiments with different configurations are presented, along with one experiment where the airbag balloon directly contacts a plate positioned on the digital scale. In the first four experiments, the contact area was assumed to be equivalent to the area covered by the donut-shaped section. In the final experiment, the contact area was defined as the area of the plate. The graph indicates that when the midsection changes its configuration, altering the contact area between the airbag balloon and the surface, Eq. (\ref{eq:forcetest}) fails to accurately estimate the applied force. This discrepancy arises because, in the current design, the airbag balloon is located beneath the midsection. While the donut-shaped midsection partially restricts the balloon, it does not significantly alter the contact area. This issue can be mitigated by positioning the donut-shaped midsection beneath the airbag balloon, allowing the midsection to directly contact the surface. When a plate with a specific area is placed beneath the airbag balloon, ensuring it is the sole contact surface, Eq. (\ref{eq:forcetest}) accurately predicts the applied force. These results indicate that while the robot does not fully adapt to varying contact areas, it can still apply consistent and precise pressure to a defined contact area.

\subsection{Testing on the casualty simulation}\label{subsec:test_casualty_simulation}

\begin{figure}[t]
\centering\includegraphics[width=\linewidth,keepaspectratio]{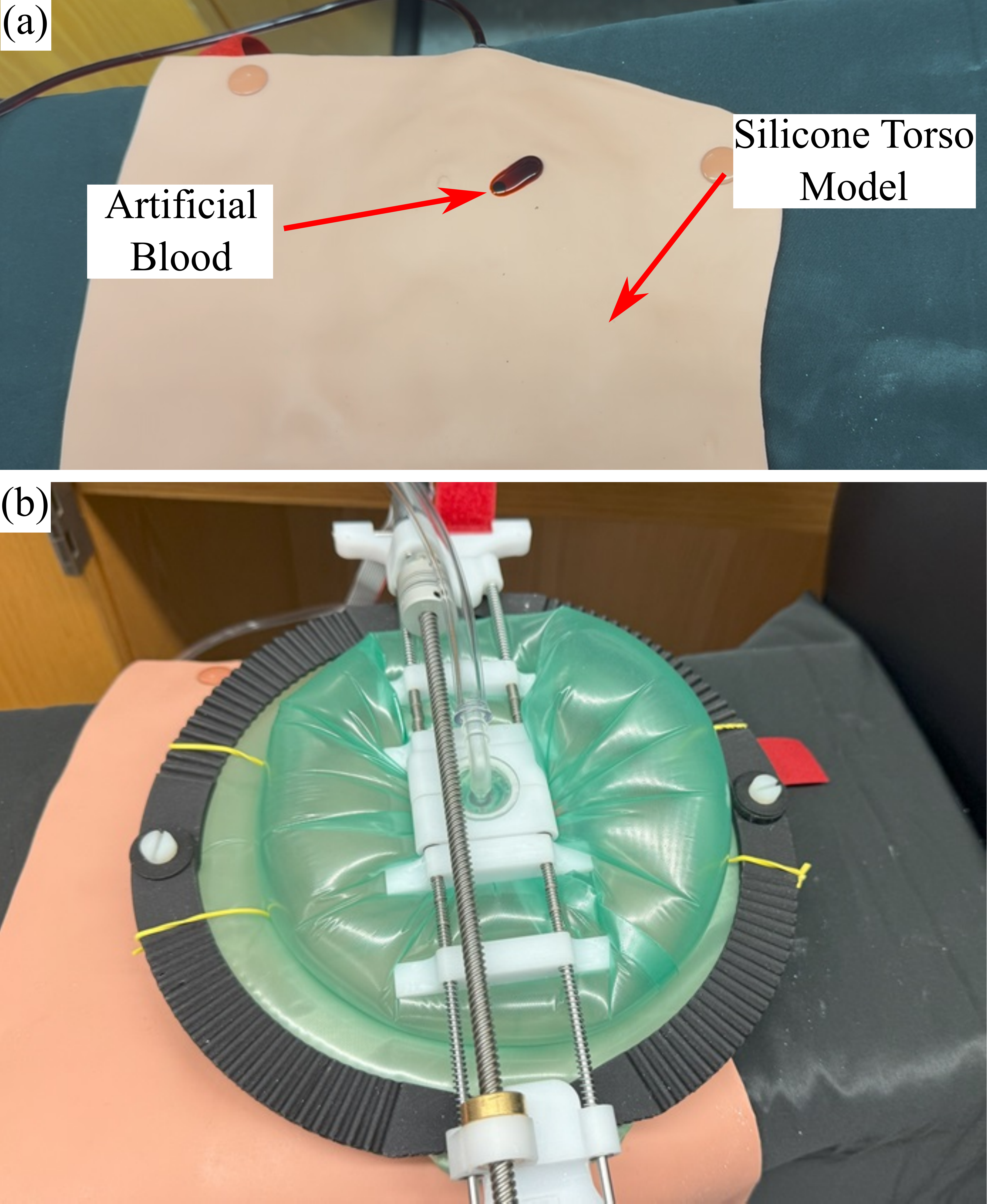}
\caption{Testing the robot on a human torso model: (a) Silicone skin with a simulation tube representing a vein, where bleeding begins to appear at a pressure of $4.83$~Kpa pressure behind the blood pump; (b) The wearable robot applied pressure to the wound area after being wrapped around the skin model, successfully stopping the bleeding. Bleeding resumed only when the pressure behind the blood pump reached $8.96$~Kpa.}
\vspace{-6 mm}
\label{fig:bleedingTest}
\end{figure}

After validating the robot on a flat surface, further experiments were performed on a human torso model to evaluate its ability to conform to the body contour and effectively mitigate bleeding. A casualty simulation kit containing artificial blood (Nasco Healthcare) was utilized, with pneumatic pumping employed to simulate blood flow through an opening on the chest, mimicking a bleeding wound (Fig. \ref{fig:bleedingTest}(a)). Blood pressure was incrementally increased until, at $4.83$~Kpa, blood flow emerged from the simulated wound. The robot was then wrapped around the torso model, and its airbag balloon was inflated with $8.27$~Kpa to exert pressure on the wound site (Fig. \ref{fig:bleedingTest}(b)). Under these conditions, the robot successfully halted the bleeding, and no blood flow was observed at the $4.83$~Kpa pressure. When the pressure of the blood pump was increased further, blood flow began to reappear at approximately $8.96$~Kpa. The results indicated that the pressure threshold for blood flow to resume with the device applied was nearly double the threshold observed in the absence of the device, confirming its efficacy in controlling bleeding.
\section{Conclusions and Future Work} \label{sec:conclusions}
In this paper, we present a wearable pressure-based robotic bleeding control device designed for managing non-extremity bleeding. The device incorporates a shape-changing ring that can transition from a circular to an ovoid configuration, enabling versatility for application to different wound shapes and sites. These ring arms could also be customized to fit the application of the device.  It features a reusable ring-arm-based actuation mechanism and extremely portable and disposable inflatable elements (an airbag balloon and inflatable ring) that are responsible for controlling the bleeding. The actuation mechanism is designed in a way that it can easily be assembled and also reused.

Furthermore, experiments were conducted to validate the robot's performance. Initially, various design configurations of the ring arm mechanism (that allows for shape changes) were tested to identify the most flexible design. Subsequently, the robot was evaluated using a digital scale to measure the force exerted by the airbag balloon as a measure of the robot's capability to apply pressure to an open wound. Finally, the robot was tested on a casualty simulation kit, where it successfully demonstrated its capability to stop simulated bleeding.
The current prototype has certain limitations: when the airbag balloon is fully inflated, the motor lacks sufficient torque to actuate the rings and modify their shape, performing effectively only when the airbag balloon is deflated or partially inflated. Furthermore, the current design does not fully conform to complex body curvatures, limiting its effectiveness in areas such as the axilla or lateral torso. Additionally, the TPE material used for the airbag balloon could benefit from a non-slip coating to enhance its ability to control bleeding more effectively. 
Future work will focus on addressing these limitations by incorporating a more powerful motor, integrating tendons within the airbag balloon to allow for direct shape manipulation, and refining the ring design to improve adaptability to various body regions. Additionally, efforts will be directed toward optimizing the overall design to make the system fully handheld, enhancing its portability and usability.
\section{Acknowledgments}
Figure \ref{fig:Configuration}(a) Created in BioRender. Yamamoto, K. (2024) https://BioRender.com/y61a492.

The material is based upon work supported by NASA Kentucky under NASA award No: 80NSSC20M0047.

\bibliographystyle{IEEEtran}
\bibliography{references}

\end{document}